\documentclass[letterpaper]{article} 
\usepackage{aaai2026}  
\usepackage{times}  
\usepackage{helvet}  
\usepackage{courier}  
\usepackage[hyphens]{url}  
\usepackage{graphicx} 
\usepackage{natbib}  
\usepackage{caption} 
\usepackage{algorithm}
\usepackage{algorithmic}
\usepackage{enumitem}
\usepackage{amssymb}
\usepackage{amsmath}
\usepackage{booktabs}
\usepackage{multicol}
\usepackage{multirow}

\usepackage{newfloat}
\usepackage{listings}
\DeclareCaptionStyle{ruled}{labelfont=normalfont,labelsep=colon,strut=off} 
\floatstyle{ruled}
\newfloat{listing}{tb}{lst}{}
\floatname{listing}{Listing}
\title{From Imitation to Discrimination: Toward A Generalized Curriculum Advantage Mechanism Enhancing Cross-Domain Reasoning Tasks}
\author {
    Changpeng Yang\textsuperscript{\rm 1,\equalcontrib}, 
    Jinyang Wu\textsuperscript{\rm 2,\equalcontrib}, 
    Yuchen Liu\textsuperscript{\rm 1}, 
    Shuai Zhang\textsuperscript{\rm 2}, 
    Yang Li\textsuperscript{\rm 1}, 
    Qiliang Liang\textsuperscript{\rm 3}, \\
    Hongzhen Wang\textsuperscript{\rm 1},
    Shuai Nie\textsuperscript{\rm 1}, 
    Jiaming Xu\textsuperscript{\rm 1,\thanks{Corresponding Author}},
    Runyu Shi\textsuperscript{\rm 1}, 
    Ying Huang\textsuperscript{\rm 1}, 
    Guoquan Zhang\textsuperscript{\rm 1}
}

\affiliations {
    \textsuperscript{\rm 1}Xiaomi Corporation,
    \textsuperscript{\rm 2}Tsinghua University,
    \textsuperscript{\rm 3}Peking University \\
}
\usepackage{bibentry}

\begin{document}

\maketitle

\begin{abstract}
Reinforcement learning has emerged as a paradigm for post-training large language models, boosting their reasoning capabilities. Such approaches compute an advantage value for each sample, reflecting better or worse performance than expected, thereby yielding both positive and negative signals for training. However, the indiscriminate mixing of the two signals in existing methods, especially from the early stages, may lead to ambiguous guidance and limited gains. To address this issue, we propose \textbf{CAPO} (\textbf{C}urriculum \textbf{A}dvantage \textbf{P}olicy \textbf{O}ptimization), an adaptive curriculum mechanism based on advantage signals. The proposed mechanism bootstraps imitation learning with positive-only advantage samples to establish robust foundations, and subsequently introduces negative signals to cultivate discriminative capabilities, thereby improving generalization across complex scenarios. Compatible with diverse optimization methods including GRPO, PPO, RLOO, and Reinforce++, our method consistently achieves stable and significant improvements in mathematical reasoning tasks, and further generalizes effectively to multimodal Graphical User Interface (GUI) reasoning scenarios, establishing itself as a versatile and robust optimization framework.
\end{abstract}


\section{Introduction}

Reinforcement learning (RL) has become a mainstream paradigm for post-training large language models, substantially advancing their reasoning capabilities, as demonstrated by DeepSeek-R1~\cite{guo2025deepseek} and Kimi-1.5~\cite{team2025kimi}. A critical component of RL algorithms such as PPO~\cite{schulman2017ppo} and GRPO~\cite{grpo} is the advantage, which quantifies whether a trajectory performs above or below expectation, providing positive and negative feedback to guide policy updates. Yet, simultaneous training on both positive and negative advantage samples often introduces ambiguity, especially during early optimization, limiting further improvement. This challenge calls for probing the essence of advantage and rethinking its role in shaping training dynamics. Since advantage inherently reflects whether the model’s competence is better or worse than expected, it naturally prompts our central research question: \textit{can advantage itself serve as a guidance indicator, enabling structured integration of positive and negative feedback into a unified, generalizable paradigm?}

\begin{figure}[t!]
\centering
\begin{minipage}{0.39\linewidth}
    \centering
    \includegraphics[width=\linewidth]{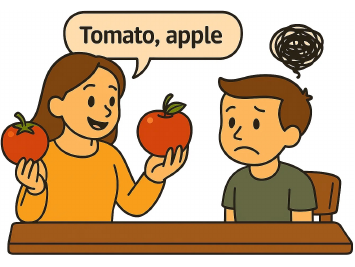}
    \textbf{(a)} RL: Mixed signals hinder learning.
\end{minipage}
\hfill
\begin{minipage}{0.59\linewidth}
    \centering
    \includegraphics[width=\linewidth]{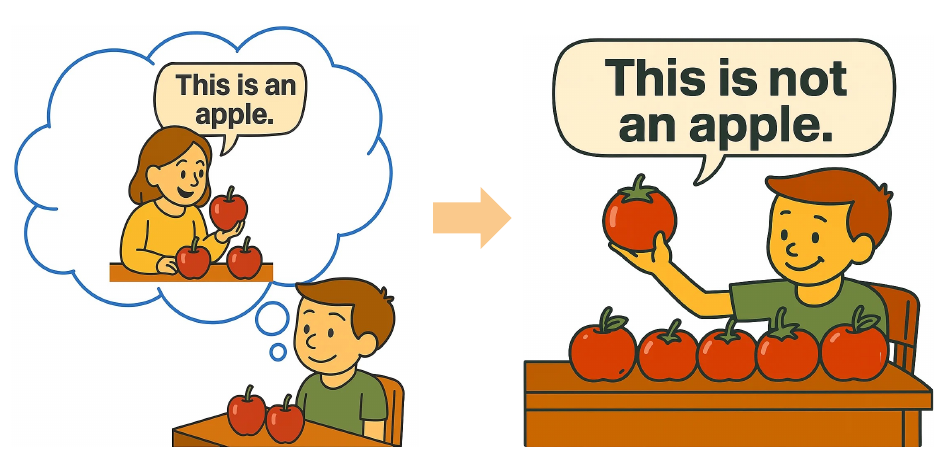}
    \textbf{(b)} CAPO: Positive imitation builds stability.
\end{minipage}
\caption{Comparison of RL and CAPO. (a) RL mixes positive and negative signals. (b) CAPO uses staged curriculum: positive imitation builds stability, negative discrimination improves generalization.}
\label{fig:Figure1}
\end{figure}

Inspired by developmental psychology, 
we note that children's learning progresses through gradual stages:
they first acquire basic behaviors through \textit{positive imitation}, 
and only later advance their generalization by incorporating \textit{corrective feedback and punishment}~\cite{bandura1961imitation,whitehurst1969discrimination}. this staged progression naturally positions advantage as an effective curriculum signal. 

Motivated by this perspective, we propose \textbf{CAPO} 
(\textbf{C}urriculum \textbf{A}dvantage \textbf{P}olicy \textbf{O}ptimization), 
a training paradigm broadly compatible with advantage-based reinforcement learning algorithms. It adopts a two-phase curriculum learning strategy based on advantage signals: in the imitation phase, positive samples foster stable foundations, and in the discriminative phase, positive samples sustain correct behaviors while negative samples refine learning, together improving generalization. As illustrated in Figure~\ref{fig:Figure1}, panel (a) shows that mixing positive and negative signals from the start leads to early confusion and prevents stable progress.  
To clarify this phenomenon, panel (b) provides a simple intuition, depicting how separating signals in stages avoids such interference.  
Together, these panels highlight the necessity of a staged curriculum design in CAPO.

While children's learning provides intuitive motivation, we provide a theoretical basis for this curriculum through the lens of the \textit{variance--bias tradeoff}. 
The imitation phase reduces gradient variance, fostering stable early training, whereas the discrimination phase restores unbiasedness, enabling effective generalization. 

We further highlight CAPO’s unique design: it leverages \textit{advantage} as a dynamic signal that aligns with the model’s evolving competence. Since curriculum learning was first introduced by Bengio et al.~\cite{bengio2009curriculum}, most methods have relied on static heuristics such as sorting tasks from easy to hard. Subsequent adaptive curriculum strategies estimate difficulty through expert annotations 
or model success rates~\cite{shi2025adaptive,chen2025sec}, 
but they remain external and heuristic, relying on manually defined proxies 
rather than signals intrinsic to the model’s evolving capbility. 
This limitation motivates our design of CAPO, which leverages advantage estimates 
as an intrinsic, competence-aware signal to drive dynamic curriculum scheduling.

Extensive experiments show that CAPO consistently enhances mathematical reasoning across diverse advantage-based RL algorithms, including GRPO, PPO, RLOO, and Reinforce++. It further demonstrates strong modality generalization, achieving notable gains on GUI-based reasoning tasks. Together, these results highlight CAPO as a versatile curriculum paradigm, effective across multiple advantage-based RL methods and transferable to multimodal reasoning domains. In summary, our main contributions are threefold:

\begin{itemize}[leftmargin=1.2em]
\item We introduce \textit{advantage} as a dynamic curriculum signal and design CAPO, a two-phase mechanism with imitation and discrimination phases adapting to the model’s evolving competence.
\item We demonstrate CAPO’s broad generalization, proving effective across diverse advantage-based RL algorithms and transferable to multimodal reasoning tasks on GUI-based environments.
\item We validate CAPO with extensive experiments, achieving improvements ranging from +2.4 to +4.0 on 1.5B models and from +1.7 to +3.9 on 7B models across diverse advantage-based RL algorithms, and also +3.81 on multimodal GUI-based reasoning tasks. 
\end{itemize}

\begin{figure*}[htbp]
\centering
\includegraphics[width=0.95\linewidth,keepaspectratio]{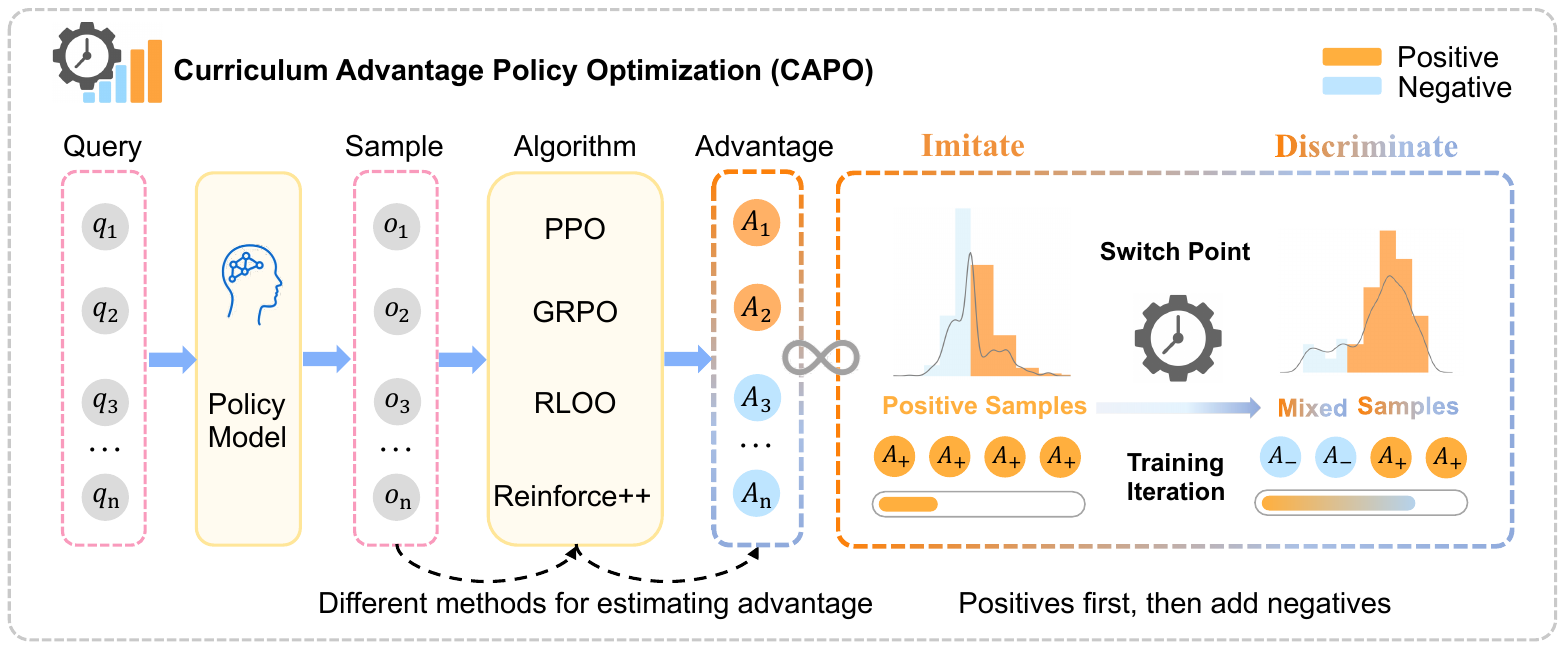}
\caption{Illustration of the CAPO scheduling mechanism. 
Each query is processed by the policy model to generate samples, 
with advantages computed under different optimization algorithms. 
In Phase~1, only positive-advantage samples are used to ensure stability; 
after the switch point, Phase~2 incorporates both positive and negative advantages 
to balance stability and generalization.}
\label{Figure2}
\end{figure*}

\section{Method}
We present CAPO, as shown in Figure~\ref{Figure2}, 
an advantage-based compatible mechanism. 
Once samples are generated by the policy model, 
various algorithms compute their respective advantage estimates, 
which are then seamlessly unified within our curriculum advantage framework. 
Leveraging these signals as guidance, CAPO explicitly structures training into 
an imitation phase and a discrimination phase, 
thereby reconciling the need for early-stage stability 
with the demand for enhanced generalization once negative samples are introduced.

\paragraph{Policy Optimization in Reinforcement Learning.}
In RL for language models, the generation process can be formulated as a policy $\pi_\theta$ that maps input prompts $\mathbf{q}$ to output sequences $\mathbf{o}$. The training objective is to maximize the expected reward assigned to the generated sequences:
\begin{equation}
\mathcal{J}(\theta) = \mathbb{E}_{\mathbf{q} \sim p_{\mathcal{Q}},\ \mathbf{o} \sim \pi_\theta(\cdot|\mathbf{q})} \left[ R(\mathbf{q}, \mathbf{o}) \right]
\label{eq:return}
\end{equation}

The policy gradient theorem~\cite{sutton1998reinforcement} 
allows reinforcement learning objectives to be optimized via gradient ascent:
\begin{equation}
\nabla_\theta \mathcal{J}(\theta) = \mathbb{E}_{\pi_\theta} 
\left[ \nabla_\theta \log \pi_\theta(\mathbf{o}|\mathbf{q}) \cdot Q^{\pi_\theta}(\mathbf{q}, \mathbf{o}) \right],
\label{eq:pg}
\end{equation}
where $Q^{\pi_\theta}(\mathbf{q}, \mathbf{o})$ denotes the action-value function, 
representing the expected return for generating sequence $\mathbf{o}$ 
from prompt $\mathbf{q}$ under policy $\pi_\theta$. 

To reduce variance in gradient estimates, it is common to replace 
$Q^{\pi_\theta}(\mathbf{q}, \mathbf{o})$ with the advantage function:
\begin{equation}
\nabla_\theta \mathcal{J}(\theta) = \mathbb{E}_{\pi_\theta} 
\left[ \nabla_\theta \log \pi_\theta(\mathbf{o}|\mathbf{q}) \cdot A^{\pi_\theta}(\mathbf{q}, \mathbf{o}) \right],
\label{eq:adv-pg}
\end{equation}
where $A^{\pi_\theta}(\mathbf{q}, \mathbf{o})$ measures whether 
a sampled trajectory outperforms the expected baseline under $\pi_\theta$. 

In practice, different policy optimization methods adopt distinct strategies 
to estimate or construct the advantage. For example, PPO employs generalized 
advantage estimation (GAE)~\cite{schulman2015gae}, which balances bias and variance 
by leveraging a discounted sum of temporal difference residuals. 
GRPO introduces a group-relative advantage formulation that normalizes advantage values 
within grouped samples to stabilize training dynamics. 
Building on these foundations, we treat the advantage as a general-purpose signal 
that not only drives gradient updates but also dynamically structures 
the training curriculum.

To stabilize training, modern methods such as PPO
optimize a clipped surrogate objective, replacing the raw policy gradient 
\begin{equation}
\mathcal{J}(\theta)
= \mathbb{E}_{\tau \sim \pi_\theta} \left[
    \frac{1}{T} \sum_{t=1}^{T}
    \min\!\left(
        \rho_t(\theta)\, A_t,\;
        \hat{\rho}_t(\theta)\, A_t
    \right)
\right],
\label{eq:clip}
\end{equation}
where $\rho_t(\theta) = \tfrac{\pi_\theta(a_t|s_t)}{\pi_{\theta_{\text{old}}}(a_t|s_t)}$ 
is the importance sampling ratio, and 
$\hat{\rho}_t(\theta) = \text{clip}(\rho_t(\theta), 1-\epsilon, 1+\epsilon)$ 
restricts updates to a stable range.

\subsection{Curriculum Advantage Policy Optimization}
We introduce \textbf{Curriculum Advantage Policy Optimization (CAPO)}, 
a progressive training framework that leverages the advantage as both a 
gradient weight and a curriculum signal. Intuitively, 
a positive advantage indicates that the model is competent on the sample, 
whereas a negative advantage suggests insufficient capability. 
Rather than applying static or heuristic curricula, CAPO dynamically 
structures learning in two stages.

\paragraph{Phase 1: Imitation Phase with Positive-Only Advantage Samples.}
Training begins with a positive-only imitation phase ($A_\tau \geq 0$), 
ensuring that updates are guided by beneficial trajectories. 
This design consolidates prior knowledge and avoids unstable gradients 
that might arise from prematurely exposing the model to challenging samples. 
Formally, building on Eq.~\ref{eq:clip} and incorporating 
a KL regularization term to prevent policy collapse 
as in RLHF~\cite{guo2025deepseek}, 
the objective is defined as:
\begin{eqnarray}
\mathcal{J}_{\text{phase-1}}(\theta)
&=& \mathbb{E}_{\tau}\Big[
\mathbb{I}_{A(\tau)\ge 0}
\Big(
\frac{1}{T}\sum_{t=1}^{T}
\min(\rho_t A_t,\hat{\rho}_t A_t)
\nonumber\\
&&\qquad\qquad
-\beta\,\mathbb{D}_{\mathrm{KL}}\big(\pi_\theta\big\|\pi_{\mathrm{ref}}\big)
\Big)
\Big].
\label{eq:phase1}
\end{eqnarray}
where the indicator $\mathbb{I}_{A_\tau \geq 0}$ filters out 
negative-advantage samples, and $\beta$ controls the KL penalty strength. 
This phase thus encourages the model to reinforce correct reasoning behaviors 
while remaining close to the reference distribution.

\paragraph{Phase 2: Discriminative Phase with Full Advantage Spectrum.}
Once a stable foundation is established, CAPO transitions into a discriminative phase 
that admits the full advantage spectrum. 
By incorporating negative-advantage samples, the model learns not only 
to reinforce strong reasoning trajectories but also to suppress suboptimal ones, 
thereby enhancing generalization. The corresponding objective is:
This progressive shift from imitation to discrimination ensures 
that CAPO first stabilizes learning and then promotes robust reasoning 
by leveraging both positive and negative feedback.

\paragraph{Curriculum Scheduling Between Phases.}
To ensure a seamless transition between the imitation and discrimination phases, 
CAPO adopts a two-stage training strategy with a predefined switch point 
(e.g., at 10\% or 20\% of total training steps). 
We also experimented with gradually introducing negative signals, 
but found that no such progressive scheme matched the effectiveness of a simple switch point. 
In practice, a hard switch provides a robust and task-agnostic mechanism that 
consistently balances early-stage stability with later discriminative learning, 
without requiring delicate hyperparameter tuning or task-specific monitoring. 
This pragmatic design ensures reproducibility across diverse settings while still 
delivering the intended variance reduction in Phase~1 and generalization gains in Phase~2.

\subsection{Theoretical Justification}
\label{sec:theory}
The CAPO algorithm’s two-phase curriculum leverages the variance--bias tradeoff in policy gradient estimation to enhance training stability and convergence. 
Consider a policy $\pi_\theta(a|s)$ with advantage estimate 
$\hat{A}(s,a) = A^\pi(s,a) + \epsilon$, 
where $\epsilon$ is zero-mean noise with variance $\sigma^2$. 
The true policy gradient $g$ and its stochastic estimate $\hat{g}$ are defined as:
\begin{align}
g &= \mathbb{E}_{\pi_\theta}\!\left[\nabla_\theta \log \pi_\theta(a|s)\, A^\pi(s,a)\right] \\
\hat{g} &= \nabla_\theta \log \pi_\theta(a|s)\, \hat{A}(s,a)
\end{align}
The mean squared error (MSE) between $\hat{g}$ and $g$ decomposes into bias and variance:
\begin{equation}
\label{eq:mse}
\mathbb{E}\!\left[\lVert \hat{g} - g \rVert^2\right] 
= \lVert \mathbb{E}[\hat{g}] - g \rVert^2 
+ \mathrm{Var}(\hat{g}).
\end{equation}

\paragraph{Phase 1 (Positive-only Imitation).} 
To suppress early-stage noise, CAPO restricts updates to positive advantages:
\begin{equation}
\hat{g}_{\text{phase-1}} 
= \mathbb{E}\!\left[\nabla_\theta \log \pi_\theta(a|s)\,
\hat{A}(s,a)\,\mathbb{I}\{\hat{A}>0\}\right].
\end{equation}
Excluding negative outliers reduces $\mathrm{Var}(\hat{g})$, 
so even with bias, the overall MSE is lowered, ensuring stable improvement.

\paragraph{Phase 2 (Full Discriminative Refinement).} 
As the policy improves and $\mathrm{Var}(\hat{A})$ shrinks, 
CAPO transitions to the unbiased estimator:
\begin{equation}
\hat{g}_{\text{phase-2}} 
= \mathbb{E}\!\left[\nabla_\theta \log \pi_\theta(a|s)\, \hat{A}(s,a)\right],
\end{equation}
restoring $\mathbb{E}[\hat{g}_{\text{phase-2}}]=g$ and enabling generalization.

\paragraph{Proposition.}
Let $\{\alpha_t\}$ denote the step sizes. Under Robbins--Monro conditions 
($\sum_t \alpha_t=\infty$, $\sum_t \alpha_t^2<\infty$), 
the CAPO update rule converges almost surely to a local optimum: 
Phase~1 bounds variance, Phase~2 restores unbiasedness, 
so the MSE in \eqref{eq:mse} vanishes asymptotically 
and the limit point of $\theta_t$ is stationary. 
A detailed proof is provided in Appendix~A.
\section{Experimental Setups}\label{experiments}

We present setups for both \textit{mathematical reasoning} 
and \textit{GUI-based multimodal} tasks, covering datasets, evaluation, and baselines.

\subsection{Mathematical Reasoning Tasks}
\paragraph{Datasets.}
Following prior works~\cite{zeng2025simplerl, liu2025understanding, wu2025thought}, we curate 5.5K Level 3–5 problems from the MATH dataset~\cite{hendrycks2021measuring}.

\paragraph{Evaluation Benchmarks.}
We evaluate on AIME 2024~\cite{li2024numinamath}, 
AMC~\cite{li2024numinamath}, 
MATH500~\cite{hendrycks2021measuring}, 
GSM8K~\cite{cobbe2021training}, 
Minerva~\cite{dataset_minerva}, 
OlympiadBench~\cite{dataset_olympiad} and  
College Math~\cite{tang2024mathscale}.  
For out-of-domain evaluation, we additionally include ARC-C~\cite{arc} and GPQA-Diamond~\cite{gpqa}.  
Inference uses greedy decoding with a 3K token budget~\cite{liu2025understanding, wu2025thought}.

\paragraph{Baselines.}
We evaluate CAPO across four mainstream reinforcement learning algorithms: GRPO, PPO, RLOO, and Reinforce++.
For each baseline, we additionally include its CAPO-augmented variant, enabling a direct assessment of CAPO’s effectiveness as a general enhancement.
This design allows us to examine both the standalone performance of the baselines and the improvements achieved through CAPO integration.

\paragraph{Implementation.}
To ensure fairness across the various baseline algorithms and CAPO variants, 
we unify the description of all implementation details. 
Because different methods require distinct parameterizations, 
the complete training configurations are provided in Appendix~B.

\subsection{GUI-based Multimodal Tasks}
\paragraph{Datasets.}
We adopt GUI-R1-3K~\cite{luo2025guir1generalistr1style}, a dataset derived from 
OS-Atlas~\cite{wu2024osatlasfoundationactionmodel}, consisting of 3K vision-language-action examples 
across web and mobile interfaces.  
The dataset pairs visual states with natural language instructions and action sequences, 
providing a compact yet diverse testbed for evaluating multimodal reasoning and planning.

\paragraph{Evaluation Benchmarks.}
\textit{Planning:} GUI-Act-Web~\cite{luo2025guir1generalistr1style}, OmniAct-Web~\cite{kapoor2024omniact}, AndroidControl-Low/High~\cite{li2024androidcontrol}.  
\textit{Perception:} Screenspot Pro~\cite{li2025screenspotpro}.

\paragraph{Baselines.}
We compare GRPO with its CAPO-augmented variant, following the dominant GUI benchmark setup~\cite{wu2024osatlasfoundationactionmodel}.  
This serves as a supplementary evaluation to demonstrate the effectiveness of CAPO in multimodal reasoning scenarios, 
beyond the primary focus on mathematical reasoning tasks.

\paragraph{Implementation.}
We train QwenVL2.5-3B\cite{qwen2.5-VL} in the VERL framework~\cite{verl2025},  
All experiments are conducted on 8×NVIDIA A100-80GB GPUs.  
To ensure reproducibility, detailed hyperparameters 
and training configurations are provided in Appendix~B.

For evaluating planning ability, we leverage four datasets across web, mobile, and desktop platforms: GUI-Act-Web\cite{luo2025guir1generalistr1style}, OmniAct-Web, AndroidControl-Low, and AndroidControl-High\cite{li2024androidcontrol, gui-odyssey}. These benchmarks involve long-horizon action prediction, decision-making under partial observability, and cross-app navigation, covering both short- and long-term planning tasks in real-world interactive environments.
We further conduct qualitative case studies on GUI-based multimodal tasks 
to illustrate the behavioral differences between baselines and CAPO-enhanced models. 
Representative examples are presented in Appendix~D.
\begin{table*}[ht!]
\centering
\renewcommand{\arraystretch}{1.15}
\setlength{\tabcolsep}{4pt}
\begin{tabular}{lcccccccc}
\hline\hline
\textbf{Method} & \textbf{AIME24$\uparrow$} & \textbf{AMC$\uparrow$} & \textbf{MATH500$\uparrow$} & \textbf{GSM8K$\uparrow$} & \textbf{Minerva$\uparrow$} & \textbf{Olympiad$\uparrow$} & \textbf{CollegeMath$\uparrow$} & \textbf{Avg.$\uparrow$}\\
\hline
\multicolumn{9}{c}{Qwen2.5-7B-Math}\\
\hline
CoT  & 13.3 & 42.5 & 50.8 & 77.8  & 22.5 & 27.8  & 42.7 & 41.3 \\
GRPO & 16.7 & 52.5 & 75.2 & 86.5 & 29.4 & 36.9 & 44.8 & 48.9 \\
GRPO (+Ours)    & \textbf{20.0} & \textbf{65.0} & \textbf{76.8} & \textbf{88.9} & \textbf{33.1} &
\textbf{39.7}  &  \textbf{46.3} & \textbf{52.8}\textsuperscript{\textbf{$\uparrow$3.9}} \\
PPO & 26.7 & 52.5 & 71.0 & 80.9 & 34.2 & 34.1 & 41.1 & 48.6\\
PPO (+Ours)   & \textbf{30.0} & \textbf{57.5} & \textbf{72.6} & \textbf{85.2} & \textbf{37.9} & \textbf{37.8} & \textbf{41.7} &  \textbf{51.8}\textsuperscript{\textbf{$\uparrow$3.2}} \\
RLOO  & 30.0 & 55.0 & 73.8 & 82.7 & 35.5 & 36.0 & 39.8 & 50.4 \\
RLOO (+Ours)  & \textbf{33.3} & \textbf{67.5} & \textbf{74.8} & \textbf{84.6} & \textbf{36.0} & 35.6 & \textbf{41.1} & \textbf{53.3}\textsuperscript{\textbf{$\uparrow$2.9}} \\
Reinforce++  & 16.7 & 52.5 & 72.4 & 85.6 & 37.1 & 37.2 & 40.3 & 48.8 \\
Reinforce++ (+Ours)  & \textbf{20.0} & \textbf{55.0} & 72.0 & \textbf{86.8} & \textbf{40.1} & 37.2 & \textbf{42.5} & 50.5\textsuperscript{\textbf{$\uparrow$1.7}} \\
\hline
\multicolumn{9}{c}{Qwen2.5-1.5B-Math}\\
\hline
CoT  & 10.0 & 42.5 & 59.0 & 74.6 & 24.3 & 27.6 & 39.5 & 39.6 \\
GRPO   & 13.3 & 52.5 & 71.2 & 83.2  & 26.8 & 30.1  & 41.2 & 45.6 \\
GRPO (+Ours)    & \textbf{23.3} & \textbf{62.5} & \textbf{71.8} & \textbf{83.9} & \textbf{32.0} &
\textbf{32.9} &  \textbf{41.7} & \textbf{49.6}\textsuperscript{\textbf{$\uparrow$4.0}}\\
PPO  & 13.3 & 50.0 & 66.6 & 74.7 & 24.6 & 27.1 & 37.8 & 42.0 \\
PPO (+Ours)  & 13.3 & \textbf{57.5} & \textbf{70.2} & \textbf{78.4} & \textbf{25.4} & \textbf{33.0} & \textbf{40.0} & \textbf{45.4}\textsuperscript{\textbf{$\uparrow$3.4}} \\
RLOO  & 20.0 & 50.0 & 68.0 & 82.6 & 28.7 & 32.0 & 41.4 & 46.1 \\
RLOO (+Ours)  & \textbf{23.3} & \textbf{57.5} & \textbf{71.6} & \textbf{83.1} & \textbf{33.8} & \textbf{33.2} & \textbf{41.8} & \textbf{49.2}\textsuperscript{\textbf{$\uparrow$3.1}}  \\
Reinforce++  & 10.0 & 47.5 & 70.0 & 83.2 & 32.0 & 31.7 & 41.3 & 45.1 \\
Reinforce++ (+Ours) & \textbf{20.0} & \textbf{50.0} & \textbf{70.8} & \textbf{83.7} & \textbf{34.2} & 31.6 & 42.0 & \textbf{47.5}\textsuperscript{\textbf{$\uparrow$2.4}} \\
\hline\hline    
\end{tabular}
\caption{We report results of different LLMs across seven mainstream benchmarks, 
with all main experiments conducted on Qwen2.5-Math-7B and Qwen2.5-Math-1.5B.  
Each baseline (GRPO, PPO, Reinforce++, and RLOO) is further evaluated with our CAPO mechanism, 
denoted as GRPO(+Ours), PPO(+Ours), and RLOO(+Ours), Reinforce++(+Ours), respectively.  
For clarity, all improvements of the CAPO variants over their corresponding baselines 
are highlighted in \textbf{bold}; identical results remain unmarked.}
\label{all_models}
\end{table*}

\begin{table*}[htbp!]
\centering
\renewcommand{\arraystretch}{1.15}
\setlength{\tabcolsep}{3pt}
\begin{tabular}{lcccccccccccccc}
\hline\hline
\textbf{Models} 
& \multicolumn{3}{c}{\textbf{GUI-Act-Web}} 
& \multicolumn{3}{c}{\textbf{OmniAct-Web}} 
& \multicolumn{3}{c}{\textbf{AndroidControl-Low}} 
& \multicolumn{3}{c}{\textbf{AndroidControl-High}} 
& \textbf{Overall} \\
\cline{2-4}\cline{5-7}\cline{8-10}\cline{11-13}
& Type & GR & SR 
& Type & GR & SR 
& Type & GR & SR 
& Type & GR & SR 
&  \\
\hline
Os-Atlas-4B 
& 79.22 & 58.57 & 42.62 
& 46.74 & 49.24 & 22.99 
& 64.58 & 71.19 & 40.62 
& 49.01 & 49.51 & 22.77 
& 49.75\\

QwenVL2.5-3B 
& 56.10 & 64.28 & 55.61 
& 50.63 & 46.89 & 47.02 
& 62.03 & 74.07 & 59.32 
& 47.81 & 46.51 & 38.90 
& 54.09\\

GRPO
& 85.10 & 82.36 & 70.23
& 79.02 & 71.10 & 70.76
& 82.13 & 80.15 & 63.87
& 60.10 & 58.25 & 46.81
& 70.79 \\

Ours
& \textbf{87.73} & \textbf{85.85} & \textbf{85.85}
& \textbf{87.24} & \textbf{74.02} & \textbf{74.16}
& \textbf{82.29} & \textbf{81.19} & 61.41
& \textbf{65.91} & \textbf{61.47} & 47.71
& \textbf{74.60} \\

$\Delta(\uparrow)$
& \textbf{$\uparrow$2.63}
& \textbf{$\uparrow$3.49}
& \textbf{$\uparrow$15.62}
& \textbf{$\uparrow$8.22}
& \textbf{$\uparrow$2.92}
& \textbf{$\uparrow$3.40}
& \textbf{$\uparrow$1.16}
& \textbf{$\uparrow$1.04}
& $\downarrow$2.46
& \textbf{$\uparrow$5.81}
& \textbf{$\uparrow$3.22}
& \textbf{$\uparrow$0.90}
& \textbf{$\uparrow$3.81} \\
\hline\hline
\end{tabular}
\caption{Performance comparison of GUI reasoning tasks across both low-level 
and high-level settings on GUI-Act-Web, OmniAct-Web, AndroidControl-Low, 
and AndroidControl-High. We report Type (action type prediction), 
GR (grounding accuracy), and SR (step success rate) under a unified 
zero-shot prompt for fair comparison. CAPO consistently improves 
over the GRPO baseline, yielding an average gain of +3.81 points 
across planning benchmarks, further demonstrating its effectiveness 
in multimodal reasoning scenarios.}
\label{tab:gui_planning}
\end{table*}

\begin{figure*}[htb]
\centering
\begin{minipage}{0.47\linewidth}
  \centering
  \includegraphics[width=\linewidth]{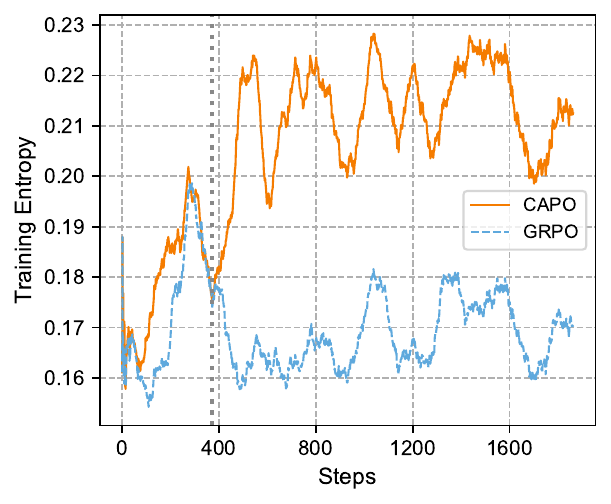}
  \caption*{(a) Entropy dynamics}
\end{minipage}\hfill
\begin{minipage}{0.47\linewidth}
  \centering
  \includegraphics[width=\linewidth]{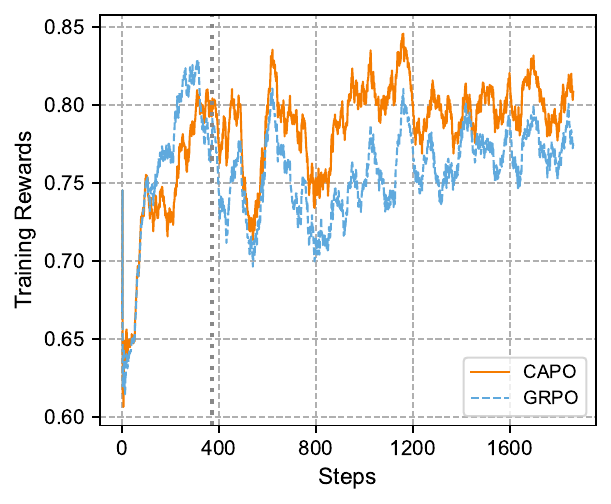}
  \caption*{(b) Reward dynamics}
\end{minipage}
\caption{Comparison of reward and entropy curves between GRPO and CAPO on the 7B model. 
The gray vertical line marks the switch from imitation to discrimination.
CAPO first relies on positive-only training to establish robust foundations. After the switch, negative samples lead to a steady increase in entropy and rewards, demonstrating enhanced generalization.}
\label{fig:reward_entropy}
\end{figure*}

\section{Experimental Results}\label{main_results}
\subsection{Main Results}
\paragraph{Math Reasoning Performance.}
Following prior works~\cite{liu2025understanding,wu2025thought}, we evaluate CAPO on seven reasoning benchmarks across two model scales and four optimization methods—GRPO~\cite{shao2024deepseekmath}, PPO~\cite{schulman2017ppo}, RLOO~\cite{ahmadian2024rloo}, and REINFORCE++~\cite{hu2025rfpp}. 

As shown in Table~\ref{all_models}, CAPO delivers consistent gains of +1.7 to +4.0 points across all methods, confirming its effectiveness as a drop-in enhancement. Notably, it achieves large improvements on competition-level tasks: for instance, AMC improves from 52.5 to 65.0 (+12.5) and AIME24 from 16.7 to 20.0 (+3.3) on the 7B model, with the 1.5B model also achieving absolute improvements of 2.4~4.0. Beyond individual datasets, consistent improvements are observed on GSM8K, OlympiadBench, Minerva, among others, demonstrating robust generalization. 

CAPO further scales well with model size: while the 7B model attains the highest performance, the 1.5B variant with CAPO closes much of the gap, approaching the larger baseline. 

Overall, CAPO shows (1) broad applicability across optimization methods, (2) scalable effectiveness that benefits both small and large models, and (3) substantial improvements on high-difficulty benchmarks, supporting advantage-based curricula as a principled mechanism to unify stability and exploration.

\paragraph{Multimodal Perception and Planning Capabilities of GUI Agent.} 
We adopt GUI-based tasks as our primary benchmark due to their complexity as cross-domain reasoning challenges in multimodal settings. These tasks require precise language understanding, fine-grained visual perception, and context integration to plan and execute actions effectively, providing a rigorous testbed for CAPO’s robustness and adaptability.

Following the setup in~\cite{luo2025guir1generalistr1style}, which employs GRPO, we evaluate performance across perception, low-level, and high-level planning tasks. As shown in Table~\ref{tab:gui_planning}, CAPO yields an overall gain of \textbf{+3.81} on planning tasks. For completeness, we also report its GUI grounding results in perception tasks, detailed in Appendix~C.

These results highlight CAPO’s strong generalization beyond mathematical reasoning, 
boosting multimodal perception and control by leveraging advantage as a curriculum signal 
that enables consistent generalization across diverse modalities.

\begin{table*}[ht!]
\centering
\renewcommand{\arraystretch}{1.15}
\begin{tabular}{lccccccc}
\hline\hline
\textbf{Method} & \textbf{AIME24$\uparrow$} & \textbf{AMC$\uparrow$} & \textbf{MATH500$\uparrow$} & \textbf{GSM8K$\uparrow$} & \textbf{Minerva$\uparrow$} & \textbf{Olympiad$\uparrow$} &\textbf{Avg.$\uparrow$}\\
\hline
CoT & 13.3 & 42.5 & 50.8 & 77.8  & 22.5 & 27.8  & 39.1 \\
ADARFT~[\citenum{li2023adaptive}] & 15.8 & 55.0 & 74.4 & \textbf{91.0} & 25.4 & 24.9 & 47.8 \\
\hline
GRPO & 16.7 & 52.5 & 75.2 & 86.5 & 29.4 & 36.9 & 49.5 \\
GRPO(+SC)& 16.7 & \textbf{65.0} & 75.0 & 86.3 & 29.8 & 38.1 & 51.8 \\
GRPO(+Ours)    & \textbf{20.0} & \textbf{65.0} & \textbf{76.8} & 88.9 & \textbf{33.1} &
\textbf{39.7}  &  \textbf{53.9} \\
\hline\hline
\end{tabular}
\caption{Performance comparison on Qwen2.5-7B-Math across six reasoning benchmarks. 
We compare five settings: CoT, ADARFT, vanilla GRPO, GRPO with a static curriculum(GRPO(+SC)),and our advantage curriculum method GRPO(+Ours). Our approach achieves the strongest overall performance among all variants.}
\label{tab:ppo-ablations}
\end{table*}

\subsection{Detailed Analysis}\label{ablation}
\begin{figure}[t]
\centering
\includegraphics[width=0.95\linewidth,keepaspectratio]{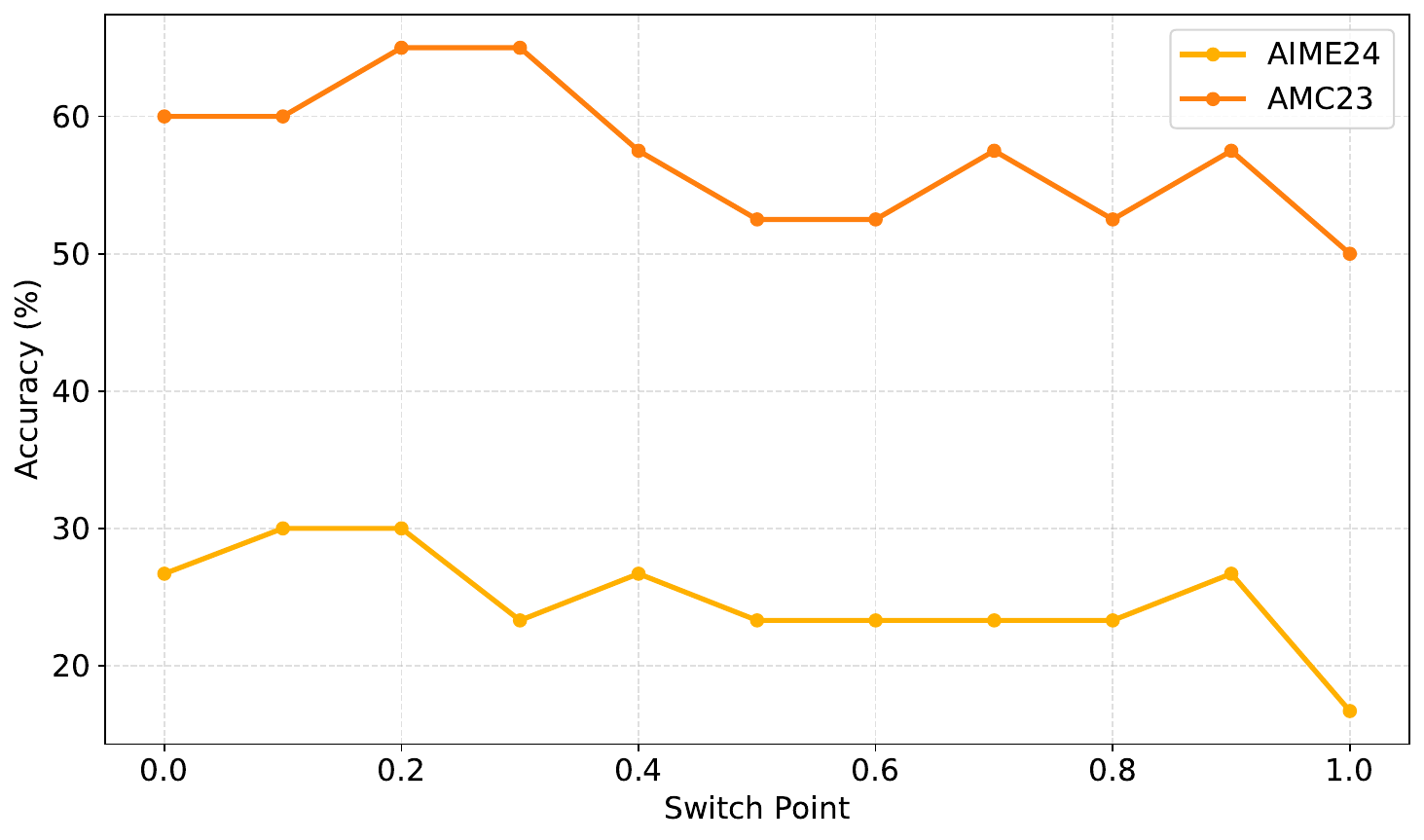}
\caption{Switch Point Sensitivity: Results on AIME24 and AMC23. While we evaluate CAPO across 8 benchmarks in total, here we present the two most representative results. The curves show that introducing the switch point in the early phase (around 0.2–0.3) yields the best performance, aligning with the theoretical expectation that early positive-only training stabilizes learning before timely inclusion of negative signals enhances generalization.}
\label{fig:switch_point}
\end{figure}

\paragraph{Training Dynamics Analysis.}\label{training_dynamics}
In Figure~\ref{fig:reward_entropy}, we present the training reward and entropy dynamics of the 7B model under CAPO and GRPO. Before the phase transition, both methods exhibit comparable reward growth; however, CAPO becomes consistently superior once imitation has stabilized. The gray vertical line marks the transition from the imitation phase to the discrimination phase. After this point, the entropy trajectory of CAPO exhibits a \textit{\underline{steady climb}}, in contrast to the plateau observed in GRPO, while rewards continue to improve. This indicates that CAPO not only secures stronger reward gains in the later stage but also maintains higher entropy, a property often linked to more diverse and exploratory reasoning paths. Moreover, the smooth rise in entropy suggests that CAPO avoids the sharp entropy collapses typically triggered by prematurely mixing negative samples. By deferring their incorporation, CAPO stabilizes the imitation stage and later exploits negative feedback more effectively, enabling better generalization across tasks.
\paragraph{Effect of Switch Stage in CAPO.}
We analyze how the switch stage influences the effectiveness of CAPO. 
Figure~\ref{fig:switch_point} reports results on two representative benchmarks (AIME24 and AMC23). 
Performance peaks when the switch occurs around 20\%–30\% of training, 
suggesting that after a short period of imitation-style learning, 
introducing negative advantages early enough encourages more discriminative reasoning 
and leads to more robust learning dynamics. 
Complete benchmark results for both  are presented in Appendix~B.1 for reference.

\paragraph{Comparing Static and Dynamic Curriculum Strategies.}
We compare the conventional static curriculum method with our dynamic advantage-based training strategy. In the static curriculum setting, we estimate sample difficulty by performing pass@16 evaluation for each sample. The dataset is then sorted based on this metric, and the model is trained following this fixed order. In contrast, our dynamic approach focuses on adjusting the advantage signal progressively during training, without the need for manually reordering the data. As shown in Table~\ref{tab:ppo-ablations}, both ADARFT and the static curriculum GRPO(+SC) offer only limited and inconsistent gains over vanilla GRPO. In contrast, our dynamic method GRPO(+Ours) achieves the strongest improvements across most tasks. This suggests that predefined difficulty heuristics are insufficient, and dynamically adjusting the advantage signal provides a more effective curriculum.

\paragraph{Further Discussion on Generalization.}\label{generalization}

\begin{figure}[htbp]
\centering
\includegraphics[width=0.95\linewidth]{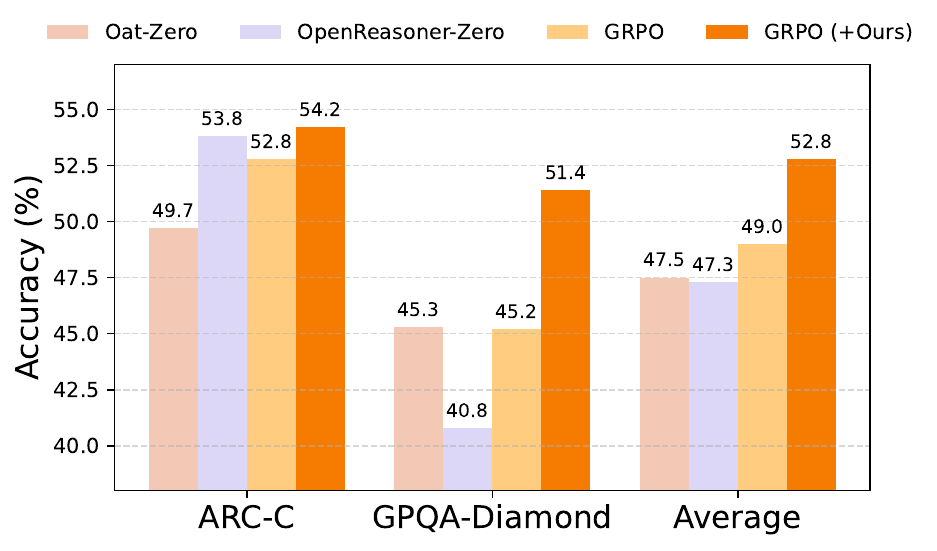}
\caption{Results on two representative out-of-distribution benchmarks 
(Qwen2.5-Math-7B-Base). CAPO achieves an average accuracy of 52.8, 
outperforming GRPO by +6.5\%, demonstrating improved robustness under distributional shifts.}
\label{ood}
\end{figure}

Recently some work has underscored the vulnerability of LLMs to distributional shifts~\cite{yuan2023revisiting,0001HH0ZWY0HGJ024}, with models often exhibiting strong in-distribution (ID) performance but significant degradation on out-of-distribution (OOD) domains~\cite{berglund2024the,yang-etal-2024-unveiling}. To evaluate CAPO’s generalization under distributional shift, we benchmark it on two representative reasoning datasets: ARC-C and GPQA-Diamond. Since all models are trained exclusively on mathematical data, this setting naturally provides a robust OOD evaluation. As shown in Figure~\ref{ood}, CAPO achieves an average accuracy of 52.8, outperforming GRPO by +3.8, with gains observed consistently across all benchmarks (+1.4 on ARC-C and +6.2 on GPQA-D). These results demonstrate the effectiveness of CAPO’s progressive learning strategy in enhancing OOD generalization by integrating both imitation and discrimination phases. These results confirm CAPO’s robustness under OOD conditions.

\section{Related Work}\label{related_works}

\paragraph{Reinforcement Learning for Large Reasoning Model.}
Recent advancements in both LLMs and MLLMs have increasingly focused on enabling models to simulate reasoning processes. Inspired by powerful reasoning models like DeepSeek-R1~\cite{guo2025deepseek}, and Kimi-k1.5~\cite{team2025kimi}, research focus has been drawn to reinforcement learning with verifiable rewards (RLVR)~\cite{wei2022chain,wang2023selfconsistency,wu2024beyond,OpenworldAUC,wu2025thought}, which combines both positive and negative feedback. Recent works have begun to explore the sperate function of them\citet{zhu2025surprising} leverage them primarily to balance diversity, yet fall short of fully unlocking their optimization potential. \citet{harnessing2025} incorporate both positive and negative samples into a DPO-inspired loss, but still rely on a fine-tuning-centric, two-stage which lacks a principled mechanism to integrate with advantage-based reinforcement learning methods.
In contrast, our method first leverages positive advantage samples to establish stable behavioral priors, then introduces negative samples to improve generalization. This staged scheduling avoids mixing signals prematurely and aligns naturally with advantage-based reinforcement learning frameworks.

\paragraph{Curriculum Learning.}
Curriculum learning (CL), originally proposed by \citet{10.1145/1553374.1553380}, has long adhered to a data-centric paradigm, and has been widely adopted in LLM development~\cite{zhu2025curicllum,xu2025towards}. Recent methods like Speed-RL~\cite{zhang2025speed} and \citet{kim2024strategic} still employ data-centric strategies that progress from simple to complex tasks using external criteria such as task diffculty to determine sample ordering. Similarly, LBS3~\cite{luo2024let} guides models through progressive training with easy-to-hard proxy queries. However, these approaches fundamentally misalign with effective curriculum design by relying on static, externally defined difficulty metrics rather than the model’s evolving capabilities. Truly effective curriculum learning should be compentece-aware, dynamically adapting to the model’s competence. 

CAPO leverages advantage estimates as an intrinsic, competence-aware signal, 
avoiding the early mixing of positive and negative feedback that often destabilizes RL training.  
It employs a staged curriculum: positive advantages first establish stable behavioral priors, 
and negative ones are later introduced to enhance generalization.  
Unlike curriculum learning methods that depend on static, externally defined difficulty measures, 
our framework adapts dynamically to the policy’s evolving competence, 
thereby unifying the stability of early reinforcement learning with the adaptability of curriculum scheduling.
\section{Conclusion}\label{conclusion}
In this paper, We propose CAPO (Curriculum Advantage Policy Optimization), a novel mechanism that addresses key limitations in reasoning model training by leveraging advantage as an intrinsic learning signal for adaptive curriculum construction. CAPO's two-phase approach progresses from imitation learning with positive-only samples to discrimination learning incorporating negative signals, mirroring human cognitive development while preventing the instability common in direct mixed-signal optimization. Extensive experiments demonstrate CAPO's consistent improvements over strong baselines like GRPO across diverse benchmarks and model scales. Remarkably, CAPO exhibits exceptional cross-domain generalization from mathematical reasoning to multimodal reasoning tasks. By aligning training with the model's evolving capabilities rather than external metrics alone, CAPO opens new avenues for developing more adaptive and cognitively-inspired learning algorithms.

\bibliography{aaai2026}

@misc{qwen2.5-VL,
    title = {Qwen2.5-VL},
    url = {https://qwenlm.github.io/blog/qwen2.5-vl/},
    author = {Qwen Team},
    month = {January},
    year = {2025}
}

@misc{li2025screenspotpro,
      title={ScreenSpot-Pro: GUI Grounding for Professional High-Resolution Computer Use}, 
      author={Kaixin Li and Ziyang Meng and Hongzhan Lin and Ziyang Luo and Yuchen Tian and Jing Ma and Zhiyong Huang and Tat-Seng Chua},
      year={2025},
      eprint={2504.07981},
      archivePrefix={arXiv},
      primaryClass={cs.CV},
      url={https://arxiv.org/abs/2504.07981}, 
}

@misc{kapoor2024omniact,
      title={OmniACT: A Dataset and Benchmark for Enabling Multimodal Generalist Autonomous Agents for Desktop and Web}, 
      author={Raghav Kapoor and Yash Parag Butala and Melisa Russak and Jing Yu Koh and Kiran Kamble and Waseem Alshikh and Ruslan Salakhutdinov},
      year={2024},
      eprint={2402.17553},
      archivePrefix={arXiv},
      primaryClass={cs.AI},
      url={https://arxiv.org/abs/2402.17553}, 
}

@misc{ahmadian2024rloo,
      title={Back to Basics: Revisiting REINFORCE Style Optimization for Learning from Human Feedback in LLMs}, 
      author={Arash Ahmadian and Chris Cremer and Matthias Gallé and Marzieh Fadaee and Julia Kreutzer and Olivier Pietquin and Ahmet Üstün and Sara Hooker},
      year={2024},
      eprint={2402.14740},
      archivePrefix={arXiv},
      primaryClass={cs.LG},
      url={https://arxiv.org/abs/2402.14740}, 
}

@misc{hu2025rfpp,
      title={REINFORCE++: An Efficient RLHF Algorithm with Robustness to Both Prompt and Reward Models}, 
      author={Jian Hu and Jason Klein Liu and Haotian Xu and Wei Shen},
      year={2025},
      eprint={2501.03262},
      archivePrefix={arXiv},
      primaryClass={cs.CL},
      url={https://arxiv.org/abs/2501.03262}, 
}

@misc{schulman2017ppo,
      title={Proximal Policy Optimization Algorithms}, 
      author={John Schulman and Filip Wolski and Prafulla Dhariwal and Alec Radford and Oleg Klimov},
      year={2017},
      eprint={1707.06347},
      archivePrefix={arXiv},
      primaryClass={cs.LG},
      url={https://arxiv.org/abs/1707.06347}, 
}

@article{schulman2015gae,
  title={High-dimensional continuous control using generalized advantage estimation},
  author={Schulman, John and Moritz, Philipp and Levine, Sergey and Jordan, Michael and Abbeel, Pieter},
  journal={arXiv preprint arXiv:1506.02438},
  year={2015}
}

@misc{chen2025sec,
      title={Self-Evolving Curriculum for LLM Reasoning}, 
      author={Xiaoyin Chen and Jiarui Lu and Minsu Kim and Dinghuai Zhang and Jian Tang and Alexandre Piché and Nicolas Gontier and Yoshua Bengio and Ehsan Kamalloo},
      year={2025},
      eprint={2505.14970},
      archivePrefix={arXiv},
      primaryClass={cs.AI},
      url={https://arxiv.org/abs/2505.14970}, 
}

@misc{shi2025adaptive,
      title={Efficient Reinforcement Finetuning via Adaptive Curriculum Learning}, 
      author={Taiwei Shi and Yiyang Wu and Linxin Song and Tianyi Zhou and Jieyu Zhao},
      year={2025},
      eprint={2504.05520},
      archivePrefix={arXiv},
      primaryClass={cs.LG},
      url={https://arxiv.org/abs/2504.05520}, 
}

@article{whitehurst1969discrimination,
  title={Discrimination learning in children as a function of reinforcement condition, task complexity, and chronological age},
  author={Whitehurst, Grover J},
  journal={Journal of Experimental Child Psychology},
  volume={7},
  number={2},
  pages={314--325},
  year={1969},
  publisher={Elsevier}
}

@article{bandura1961imitation,
  title={Transmission of aggression through imitation of aggressive models.},
  author={Bandura, Albert and Ross, Dorothea and Ross, Sheila A},
  journal={The Journal of Abnormal and Social Psychology},
  volume={63},
  number={3},
  pages={575},
  year={1961},
  publisher={American Psychological Association}
}

@inproceedings{bengio2009curriculum,
  title={Curriculum learning},
  author={Bengio, Yoshua and Louradour, J{\'e}r{\^o}me and Collobert, Ronan and Weston, Jason},
  booktitle={Proceedings of the 26th annual international conference on machine learning},
  pages={41--48},
  year={2009}
}

@article{li2023adaptive,
  title={Deep reinforcement learning for adaptive learning systems},
  author={Li, Xiao and Xu, Hanchen and Zhang, Jinming and Chang, Hua-hua},
  journal={Journal of Educational and Behavioral Statistics},
  volume={48},
  number={2},
  pages={220--243},
  year={2023},
  publisher={SAGE Publications Sage CA: Los Angeles, CA}
}

@inproceedings{verl2025, series={EuroSys ’25},
   title={HybridFlow: A Flexible and Efficient RLHF Framework},
   url={http://dx.doi.org/10.1145/3689031.3696075},
   DOI={10.1145/3689031.3696075},
   booktitle={Proceedings of the Twentieth European Conference on Computer Systems},
   publisher={ACM},
   author={Sheng, Guangming and Zhang, Chi and Ye, Zilingfeng and Wu, Xibin and Zhang, Wang and Zhang, Ru and Peng, Yanghua and Lin, Haibin and Wu, Chuan},
   year={2025},
   month=mar, pages={1279–1297},
   collection={EuroSys ’25} }

@misc{wu2024osatlasfoundationactionmodel,
      title={OS-ATLAS: A Foundation Action Model for Generalist GUI Agents}, 
      author={Zhiyong Wu and Zhenyu Wu and Fangzhi Xu and Yian Wang and Qiushi Sun and Chengyou Jia and Kanzhi Cheng and Zichen Ding and Liheng Chen and Paul Pu Liang and Yu Qiao},
      year={2024},
      eprint={2410.23218},
      archivePrefix={arXiv},
      primaryClass={cs.CL},
      url={https://arxiv.org/abs/2410.23218}, 
}

@misc{gui-odyssey,
      title={GUI Odyssey: A Comprehensive Dataset for Cross-App GUI Navigation on Mobile Devices}, 
      author={Quanfeng Lu and Wenqi Shao and Zitao Liu and Fanqing Meng and Boxuan Li and Botong Chen and Siyuan Huang and Kaipeng Zhang and Yu Qiao and Ping Luo},
      year={2024},
      eprint={2406.08451},
      archivePrefix={arXiv},
      primaryClass={cs.CV},
      url={https://arxiv.org/abs/2406.08451}, 
}

@misc{li2024androidcontrol,
      title={On the Effects of Data Scale on UI Control Agents}, 
      author={Wei Li and William Bishop and Alice Li and Chris Rawles and Folawiyo Campbell-Ajala and Divya Tyamagundlu and Oriana Riva},
      year={2024},
      eprint={2406.03679},
      archivePrefix={arXiv},
      primaryClass={cs.AI},
      url={https://arxiv.org/abs/2406.03679}, 
}

@misc{luo2025guir1generalistr1style,
      title={GUI-R1 : A Generalist R1-Style Vision-Language Action Model For GUI Agents}, 
      author={Run Luo and Lu Wang and Wanwei He and Xiaobo Xia},
      year={2025},
      eprint={2504.10458},
      archivePrefix={arXiv},
      primaryClass={cs.CV},
      url={https://arxiv.org/abs/2504.10458}, 
}

@article{luo2024let,
  title={Let's Be Self-generated via Step by Step: A Curriculum Learning Approach to Automated Reasoning with Large Language Models},
  author={Luo, Kangyang and Ding, Zichen and Weng, Zhenmin and Qiao, Lingfeng and Zhao, Meng and Li, Xiang and Yin, Di and Shu, Jinlong},
  journal={arXiv preprint arXiv:2410.21728},
  year={2024}
}

@article{kim2024strategic,
  title={Strategic Data Ordering: Enhancing Large Language Model Performance through Curriculum Learning},
  author={Kim, Jisu and Lee, Juhwan},
  journal={arXiv preprint arXiv:2405.07490},
  year={2024}
}

@article{zhang2025speed,
  title={SPEED-RL: Faster Training of Reasoning Models via Online Curriculum Learning},
  author={Zhang, Ruiqi and Arora, Daman and Mei, Song and Zanette, Andrea},
  journal={arXiv preprint arXiv:2506.09016},
  year={2025}
}

@inproceedings{10.1145/1553374.1553380,
author = {Bengio, Yoshua and Louradour, J\'{e}r\^{o}me and Collobert, Ronan and Weston, Jason},
title = {Curriculum learning},
year = {2009},
isbn = {9781605585161},
publisher = {Association for Computing Machinery},
address = {New York, NY, USA},
booktitle = {Proceedings of the 26th Annual International Conference on Machine Learning},
pages = {41–48},
numpages = {8},
location = {Montreal, Quebec, Canada},
series = {ICML '09}
}

@article{wei2022chain,
  title={Chain-of-thought prompting elicits reasoning in large language models},
  author={Wei, Jason and Wang, Xuezhi and Schuurmans, Dale and Bosma, Maarten and Xia, Fei and Chi, Ed and Le, Quoc V and Zhou, Denny and others},
  journal={Advances in neural information processing systems},
  volume={35},
  pages={24824--24837},
  year={2022}
}

@article{guo2025deepseek,
  title={Deepseek-r1: Incentivizing reasoning capability in llms via reinforcement learning},
  author={Guo, Daya and Yang, Dejian and Zhang, Haowei and Song, Junxiao and Zhang, Ruoyu and Xu, Runxin and Zhu, Qihao and Ma, Shirong and Wang, Peiyi and Bi, Xiao and others},
  journal={arXiv preprint arXiv:2501.12948},
  year={2025}
}

@article{shao2024deepseekmath,
  title={Deepseekmath: Pushing the limits of mathematical reasoning in open language models},
  author={Shao, Zhihong and Wang, Peiyi and Zhu, Qihao and Xu, Runxin and Song, Junxiao and Bi, Xiao and Zhang, Haowei and Zhang, Mingchuan and Li, YK and Wu, Y and others},
  journal={arXiv preprint arXiv:2402.03300},
  year={2024}
}

@book{sutton1998reinforcement,
  title={Reinforcement learning: An introduction},
  author={Sutton, Richard S and Barto, Andrew G and others},
  volume={1},
  number={1},
  year={1998},
  publisher={MIT press Cambridge}
}

@article{arc,
      author    = {Peter Clark  and Isaac Cowhey and Oren Etzioni and Tushar Khot and
                    Ashish Sabharwal and Carissa Schoenick and Oyvind Tafjord},
      title     = {Think you have Solved Question Answering? Try ARC, the AI2 Reasoning Challenge},
      journal   = {arXiv:1803.05457v1},
      year      = {2018},
}

@inproceedings{gpqa,
      title={{GPQA}: A Graduate-Level Google-Proof Q\&A Benchmark},
      author={David Rein and Betty Li Hou and Asa Cooper Stickland and Jackson Petty and Richard Yuanzhe Pang and Julien Dirani and Julian Michael and Samuel R. Bowman},
      booktitle={First Conference on Language Modeling},
      year={2024},
      url={https://openreview.net/forum?id=Ti67584b98}
}

@article{cobbe2021training,
  title={Training verifiers to solve math word problems},
  author={Cobbe, Karl and Kosaraju, Vineet and Bavarian, Mohammad and Chen, Mark and Jun, Heewoo and Kaiser, Lukasz and Plappert, Matthias and Tworek, Jerry and Hilton, Jacob and Nakano, Reiichiro and others},
  journal={arXiv preprint arXiv:2110.14168},
  year={2021}
}

@inproceedings{dataset_olympiad,
  title={OlympiadBench: A Challenging Benchmark for Promoting AGI with Olympiad-Level Bilingual Multimodal Scientific Problems},
  author={He, Chaoqun and Luo, Renjie and Bai, Yuzhuo and Hu, Shengding and Thai, Zhen and Shen, Junhao and Hu, Jinyi and Han, Xu and Huang, Yujie and Zhang, Yuxiang and others},
  booktitle={Proceedings of the 62nd Annual Meeting of the Association for Computational Linguistics (Volume 1: Long Papers)},
  pages={3828--3850},
  year={2024}
}

@article{dataset_minerva,
  title={Solving quantitative reasoning problems with language models},
  author={Lewkowycz, Aitor and Andreassen, Anders and Dohan, David and Dyer, Ethan and Michalewski, Henryk and Ramasesh, Vinay and Slone, Ambrose and Anil, Cem and Schlag, Imanol and Gutman-Solo, Theo and others},
  journal={Advances in Neural Information Processing Systems},
  volume={35},
  pages={3843--3857},
  year={2022}
}

@misc{li2024numinamath,
  author       = {Jia Li and Edward Beeching and Lewis Tunstall and Ben Lipkin and Roman Soletskyi and Shengyi Huang and Kashif Rasul and Longhui Yu and Albert Q. Jiang and Ziju Shen and others},
  title        = {Numinamath: The largest public dataset in AI4Maths with 860k pairs of competition math problems and solutions},
  year         = {2024},
  howpublished = {\url{https://huggingface.co/datasets/Numinamath}},
  note         = {Hugging Face repository, 13:9}
}

@article{zeng2025simplerl,
  title={Simplerl-zoo: Investigating and taming zero reinforcement learning for open base models in the wild},
  author={Zeng, Weihao and Huang, Yuzhen and Liu, Qian and Liu, Wei and He, Keqing and Ma, Zejun and He, Junxian},
  journal={arXiv preprint arXiv:2503.18892},
  year={2025}
}

@inproceedings{
hendrycks2021measuring,
title={Measuring Mathematical Problem Solving With the {MATH} Dataset},
author={Dan Hendrycks and Collin Burns and Saurav Kadavath and Akul Arora and Steven Basart and Eric Tang and Dawn Song and Jacob Steinhardt},
booktitle={Thirty-fifth Conference on Neural Information Processing Systems Datasets and Benchmarks Track (Round 2)},
year={2021}
}

@article{liu2025understanding,
  title={Understanding r1-zero-like training: A critical perspective},
  author={Liu, Zichen and Chen, Changyu and Li, Wenjun and Qi, Penghui and Pang, Tianyu and Du, Chao and Lee, Wee Sun and Lin, Min},
  journal={arXiv preprint arXiv:2503.20783},
  year={2025}
}

@article{team2025kimi,
  title={Kimi k1. 5: Scaling reinforcement learning with llms},
  author={Team, Kimi and Du, Angang and Gao, Bofei and Xing, Bowei and Jiang, Changjiu and Chen, Cheng and Li, Cheng and Xiao, Chenjun and Du, Chenzhuang and Liao, Chonghua and others},
  journal={arXiv preprint arXiv:2501.12599},
  year={2025}
}

@article{xu2025towards,
  title={Towards Large Reasoning Models: A Survey of Reinforced Reasoning with Large Language Models},
  author={Xu, Fengli and Hao, Qianyue and Zong, Zefang and Wang, Jingwei and Zhang, Yunke and Wang, Jingyi and Lan, Xiaochong and Gong, Jiahui and Ouyang, Tianjian and Meng, Fanjin and others},
  journal={arXiv preprint arXiv:2501.09686},
  year={2025}
}

@inproceedings{
wang2023selfconsistency,
title={Self-Consistency Improves Chain of Thought Reasoning in Language Models},
author={Xuezhi Wang and Jason Wei and Dale Schuurmans and Quoc V Le and Ed H. Chi and Sharan Narang and Aakanksha Chowdhery and Denny Zhou},
booktitle={The Eleventh International Conference on Learning Representations },
year={2023}
}

@misc{grpo,
      title={DeepSeekMath: Pushing the Limits of Mathematical Reasoning in Open Language Models}, 
      author={Zhihong Shao and Peiyi Wang and Qihao Zhu and Runxin Xu and Junxiao Song and Xiao Bi and Haowei Zhang and Mingchuan Zhang and Y. K. Li and Y. Wu and Daya Guo},
      year={2024},
      eprint={2402.03300},
      archivePrefix={arXiv},
      primaryClass={cs.CL},
      url={https://arxiv.org/abs/2402.03300}, 
}

@article{ppo,
  title={Proximal policy optimization algorithms},
  author={Schulman, John and Wolski, Filip and Dhariwal, Prafulla and Radford, Alec and Klimov, Oleg},
  journal={arXiv preprint arXiv:1707.06347},
  year={2017}
}

@inproceedings{yang-etal-2024-unveiling,
    title = "Unveiling the Generalization Power of Fine-Tuned Large Language Models",
    author = "Yang, Haoran  and
      Zhang, Yumeng  and
      Xu, Jiaqi  and
      Lu, Hongyuan  and
      Heng, Pheng-Ann  and
      Lam, Wai",
    booktitle = "Proceedings of the 2024 Conference of the North American Chapter of the Association for Computational Linguistics: Human Language Technologies (Volume 1: Long Papers)",
    month = jun,
    year = "2024",
    address = "Mexico City, Mexico",
    publisher = "Association for Computational Linguistics",
    pages = "884--899"
}

@inproceedings{
berglund2024the,
title={The Reversal Curse: {LLM}s trained on {\textquotedblleft}A is B{\textquotedblright} fail to learn {\textquotedblleft}B is A{\textquotedblright}},
author={Lukas Berglund and Meg Tong and Maximilian Kaufmann and Mikita Balesni and Asa Cooper Stickland and Tomasz Korbak and Owain Evans},
booktitle={The Twelfth International Conference on Learning Representations},
year={2024}
}

@article{0001HH0ZWY0HGJ024,
  author={Jindong Wang and Xixu Hu and Wenxin Hou and Hao Chen and Runkai Zheng and Yidong Wang and Linyi Yang and Wei Ye and Haojun Huang and Xiubo Geng and Binxing Jiao and Yue Zhang and Xing Xie},
  title={On the Robustness of ChatGPT: An Adversarial and Out-of-distribution Perspective},
  year={2024},
  cdate={1704067200000},
  journal={IEEE Data Eng. Bull.},
  volume={47},
  number={1},
  pages={48-62}
}

@inproceedings{
yuan2023revisiting,
title={Revisiting Out-of-distribution Robustness in {NLP}: Benchmarks, Analysis, and {LLM}s Evaluations},
author={Lifan Yuan and Yangyi Chen and Ganqu Cui and Hongcheng Gao and FangYuan Zou and Xingyi Cheng and Heng Ji and Zhiyuan Liu and Maosong Sun},
booktitle={Thirty-seventh Conference on Neural Information Processing Systems Datasets and Benchmarks Track},
year={2023}
}

@article{OpenworldAUC,
  author       = {Cong Hua and
                  Qianqian Xu and
                  Zhiyong Yang and
                  Zitai Wang and
                  Shilong Bao and
                  Qingming Huang},
  title        = {OpenworldAUC: Towards Unified Evaluation and Optimization for Open-world
                  Prompt Tuning},
  journal      = {CoRR},
  volume       = {abs/2505.05180},
  year         = {2025}
}

@article{wu2024beyond,
  title={Beyond examples: High-level automated reasoning paradigm in in-context learning via mcts},
  author={Wu, Jinyang and Feng, Mingkuan and Zhang, Shuai and Che, Feihu and Wen, Zengqi and Tao, Jianhua},
  journal={arXiv preprint arXiv:2411.18478},
  year={2024}
}

@article{wu2025thought,
  title={Thought-Augmented Policy Optimization: Bridging External Guidance and Internal Capabilities},
  author={Wu, Jinyang and Liao, Chonghua and Feng, Mingkuan and Zhang, Shuai and Wen, Zhengqi and Shao, Pengpeng and Xu, Huazhe and Tao, Jianhua},
  journal={arXiv preprint arXiv:2505.15692},
  year={2025}
}

@article{harnessing2025,
  title={Harnessing Negative Signals: Reinforcement Distillation from Teacher Data for LLM Reasoning},
  author={Xu, Shuyao and Peng, Cheng and Long, Jiangxuan and Xu, Weidi and Chu, Wei and Qi, Yuan},
  journal={arXiv preprint arXiv:2505.24850},
  year={2025}
}

@article{zhu2025surprising,
  title={The Surprising Effectiveness of Negative Reinforcement in LLM Reasoning},
  author={Zhu, Xinyu and Xia, Mengzhou and Wei, Zhepei and Chen, Wei-Lin and Chen, Danqi and Meng, Yu},
  journal={arXiv preprint arXiv:2506.01347},
  year={2025}
}

@article{zhu2025curicllum,
  title={Curriculum Reinforcement Learning from Easy to Hard Tasks Improves LLM Reasoning},
  author={Shubham Parashar and Shurui Gui and Xiner Li and Hongyi Ling and Sushil Vemuri and Blake Olson and Eric Li and Yu Zhang and James Caverlee and Dileep Kalathil and Shuiwang Ji},
  journal={arXiv preprint arXiv:2506.06632},
  year={2025}
}

@article{tang2024mathscale,
  title={Mathscale: Scaling instruction tuning for mathematical reasoning},
  author={Tang, Zhengyang and Zhang, Xingxing and Wang, Benyou and Wei, Furu},
  journal={arXiv preprint arXiv:2403.02884},
  year={2024}
}

\newpage
\onecolumn
\appendix
\section{Theoretical Derivations}
\label{app:theory}

In this appendix, we provide detailed derivations of the 
variance--bias tradeoff and convergence guarantees for CAPO, 
extending the theoretical motivation outlined in the main paper.

Consider the policy gradient estimate
\[
\hat{g} = \nabla_\theta \log \pi_\theta(a|s)\, \hat{A}(s,a),
\]
where
\[
\hat{A}(s,a) = A^\pi(s,a) + \epsilon, \quad 
\mathbb{E}[\epsilon] = 0, \quad 
\mathrm{Var}[\epsilon] = \sigma^2.
\]
We assume $\epsilon$ is independent across state--action pairs and bounded, 
ensuring finite moments.  
Let the true gradient be
\[
g = \mathbb{E}\big[\nabla_\theta \log \pi_\theta(a|s)\, A^\pi(s,a)\big].
\]

The mean squared error (MSE) of $\hat{g}$ is:
\[
\mathbb{E}\big[\lVert \hat{g} - g \rVert^2\big] 
= \mathrm{Var}(\hat{g}) 
+ \lVert \mathbb{E}[\hat{g}] - g \rVert^2,
\]
since $\mathbb{E}[\hat{g} - \mathbb{E}[\hat{g}]] = 0$.  
If $\mathbb{E}[\hat{A}(s,a)] = A^\pi(s,a)$, then $\mathbb{E}[\hat{g}] = g$, 
so the bias term vanishes.  
The variance becomes:
\[
\mathrm{Var}(\hat{g}) 
= \mathbb{E}\big[\lVert \nabla_\theta \log \pi_\theta(a|s)\, 
\hat{A}(s,a) \rVert^2\big] - \lVert g \rVert^2.
\]
Large variance in $\hat{A}(s,a)$ inflates $\mathrm{Var}(\hat{g})$, 
destabilizing training.

\subsection{Phase 1: Variance Reduction}

In Phase 1, CAPO filters updates to only positive advantages:
\[
\hat{g}_{\text{phase-1}} 
= \mathbb{E}\!\left[\nabla_\theta \log \pi_\theta(a|s)\,
\hat{A}(s,a)\, \mathbb{1}\{\hat{A} > 0\}\right].
\]

Its expectation is:
\[
\mathbb{E}[\hat{g}_{\text{phase-1}}] 
= \mathbb{E}\!\left[\nabla_\theta \log \pi_\theta(a|s)\,
\hat{A}(s,a)\, \mathbb{1}\{\hat{A} > 0\}\right] \neq g,
\]
indicating bias unless $\mathbb{P}(\hat{A} > 0) = 1$ 
or the negative advantages have no net effect.  
The bias term is:
\[
\mathbb{E}[\hat{g}_{\text{phase-1}}] - g 
= \mathbb{E}\!\left[\nabla_\theta \log \pi_\theta(a|s)\,
\hat{A}(s,a)\, \mathbb{1}\{\hat{A} \leq 0\}\right].
\]

The variance is:
\[
\mathrm{Var}(\hat{g}_{\text{phase-1}}) 
= \mathbb{E}\!\left[\lVert \nabla_\theta \log \pi_\theta(a|s)\,
\hat{A}(s,a)\, \mathbb{1}\{\hat{A} > 0\} \rVert^2\right] 
- \lVert \mathbb{E}[\hat{g}_{\text{phase-1}}] \rVert^2.
\]

Since $\mathbb{1}\{\hat{A} > 0\}$ filters out negative samples—
often high-magnitude in early training—variance is reduced.  
For example, if $\hat{A} \sim \mathcal{N}(0, \sigma^2)$, 
symmetry implies filtering halves the variance contribution.

Although bias is introduced, it is tolerable early on, 
as it steers the policy toward high-reward actions, 
thus reducing $\mathrm{Var}(\hat{A}\mid \pi_\theta)$.

\subsection{Phase 2: Convergence Guarantee}

As the policy $\pi_\theta$ improves, 
$\mathrm{Var}(\hat{A}\mid \pi_\theta)$ decreases.  
CAPO then switches to the unbiased estimator:
\[
\hat{g}_{\text{phase-2}} 
= \mathbb{E}\!\left[\nabla_\theta \log \pi_\theta(a|s)\,
\hat{A}(s,a)\right],
\]
with variance
\[
\mathrm{Var}(\hat{g}_{\text{phase-2}}) 
= \mathbb{E}\!\left[\lVert \nabla_\theta \log \pi_\theta(a|s)\,
\hat{A}(s,a) \rVert^2\right] - \lVert g \rVert^2.
\]

Given bounded $\nabla_\theta \log \pi_\theta(a|s)$ 
and $\hat{A}(s,a)$, and under Robbins–Monro conditions
($\sum_t \alpha_t = \infty$, $\sum_t \alpha_t^2 < \infty$),
stochastic gradient descent converges almost surely 
to a stationary point of the expected reward.

\section{Experimental Settings and Hyperparameters} 
We provide detailed experimental settings and hyperparameters to ensure reproducibility.The experiments are divided into two parts: mathematical reasoning and multimodla Graphics User Interface(GUI) tasks.
\subsection{Mathematical Reasoning Experiments}
\paragraph{Training Setup.}
We use Qwen2.5-Math Models at two scales(1.5B and 7B). The maximum input length is set to 1024 tokens and max completion length set to 1024 tokens, All experiments are trained with the AdamW optimizer.
The initial learning rate is $1\times 10^{-5}$, for GRPO algorithms, we generate 16 candidates for each prompt and train epochs to 2. 
For algorithms as PPO, RLOO and Reinforce++, we set the batch size as 128, we use the parameter switch point to determine the time point to introduce the negative samples. we use temperature = 0.7 and top-$p$ = 0.7.

\subsection{GUI reasoning tasks}

\paragraph{Training Setup.}
We use Qwen2.5-VL-3B-Instruct as the base model. he maximum input length is set to 1024 tokens, and the maximum completion length is also set to 1024 tokens. The AdamW optimizer is used with a learning rate of $5\times10^{-6}$ 
and a cosine decay schedule. 
Batch size is 128. 
Training runs for 3 epochs with gradient accumulation steps set to 8. 
KL regularization is set to 0.02, and dropout is applied with probability 0.1. 
Mixed precision (bfloat16) is enabled.

\paragraph{Reward and Advantage.}
GUI grounding rewards are based on ScreenSpot-Pro annotations. 
We compute token-level group relative advantage 
and update the policy using $\mu=2$ GRPO iterations per batch. 
CAPO employs a hard switch at 30\% of training steps by default. We use a combination of two rewards: the coordinate reward, which equals 1 if the predicted point lies within the bounding box and 0 otherwise, and the format reward, which is granted when the output sequence conforms to the required format specified in the prompt.

\section{Comprehensive Experimental Results}
\subsection{Full Results of Switch Stage Sensitivity Study}
\begin{figure}[htbp]
    \centering
    \includegraphics[width=1.0\linewidth]{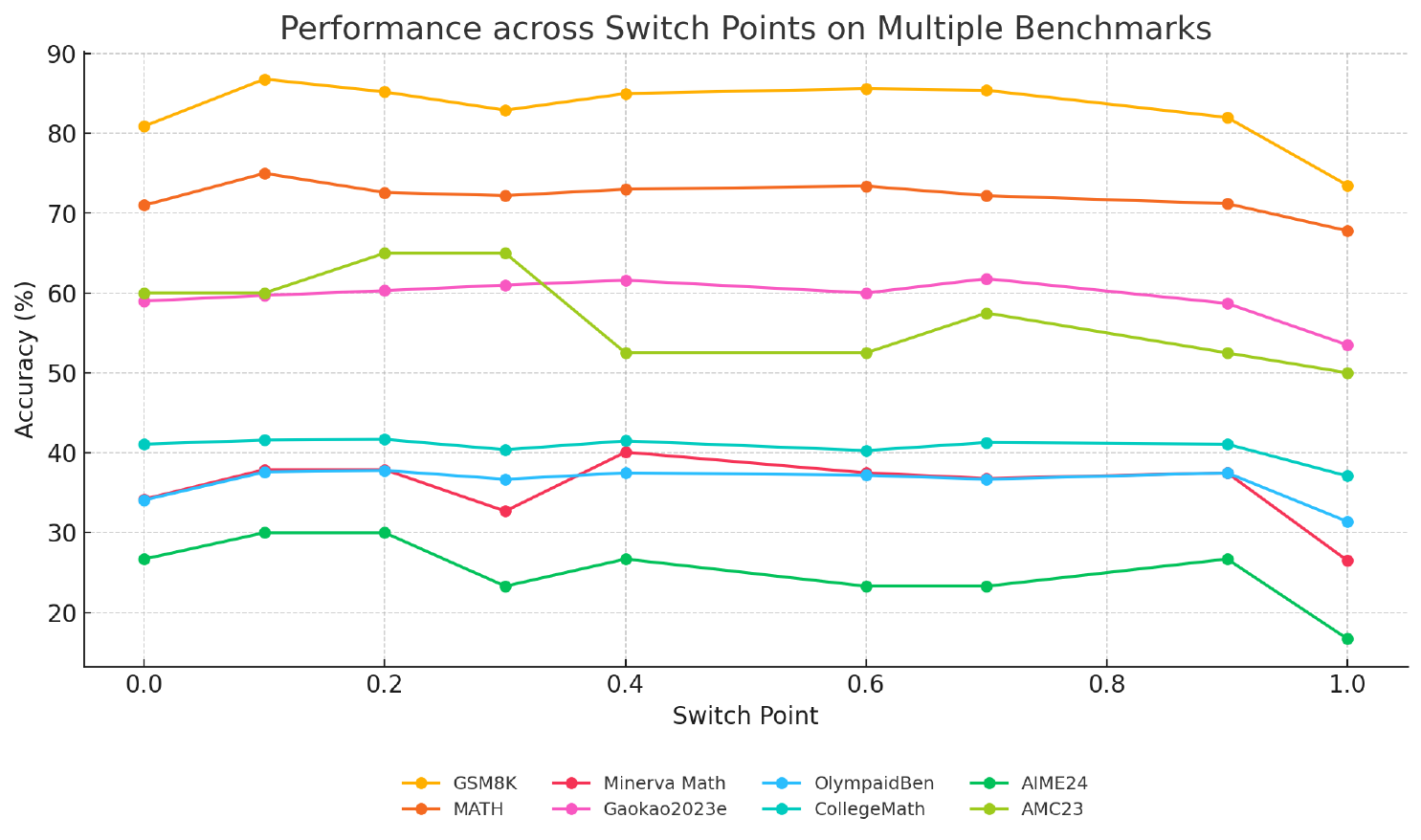}
    \caption{We present results on eight benchmarks to examine the influence of the switch point for introducing negative samples. As shown in Figure~\ref{fig:abliation_switch}, AIME24, AMC23, GSM8K and MATH exhibit a clear performance increase when the switch occurs in the early stage(Approximately 10\%-40\%, whereas College Math and OlmpaidBench display a more stable trend with moderate fluctuations.)}
    \label{fig:abliation_switch}
\end{figure}

\subsection{Results on GUI-based Perception results}
Table~\ref{tab:gui_perception} presents the GUI grounding accuracy results on the ScreenSpot-Pro benchmark across four domains: Dev, Creative, CAD, and Scientific. Compared with strong baselines such as CogAgent-18B, UGround-7B, and Os-Atlas-7B, our method consistently improves grounding accuracy, particularly on text-based grounding tasks. For example, in the Creative domain, our approach achieves 37.37 in text accuracy and 4.19 in icon accuracy, outperforming GRPO by +3.54 and +3.59 points, respectively. Similar gains are observed in the CAD and Scientific domains, where improvements reach up to +7.11 and +6.94 points.

These results highlight the effectiveness of our proposed CAPO strategy in enhancing grounding performance across diverse domains. By adaptively balancing stability and exploration, our approach yields consistent and substantial improvements, demonstrating strong generalization capability in both text and icon grounding tasks.
\begin{table}[hbp] 
\centering
\begin{tabular}{lcc cc cc cc}
\toprule
\multirow{3}{*}{Models} &  \multicolumn{8}{c}{ScreenSpot-Pro} \\
 & \multicolumn{2}{c}{Dev} & \multicolumn{2}{c}{Creative} & \multicolumn{2}{c}{CAD} & \multicolumn{2}{c}{Scientific} \\
 \cmidrule(lr){2-3}
 \cmidrule(lr){4-5}
 \cmidrule(lr){6-7}
 \cmidrule(lr){8-9}
  & Text & Icon & Text & Icon & Text & Icon & Text & Icon \\
\midrule
CogAgent-18B & 14.9 & 0.7 & 9.6 & 0.0 & 7.1 & 3.1 & 22.2 & 1.8 \\
UGround-7B   & 26.6 & 2.1 & 27.3 & 2.8 & 14.2 & 1.6 & 31.9 & 2.7 \\
Os-Atlas-7B  & 33.1 & 1.4 & 28.8 & 2.8 & 12.2 & 4.7 & 37.5 & 7.3 \\
\midrule
GRPO & 29.22 & 4.13 & 33.83 & 0.60  & 25.88 & 6.25 & 50.00 & 18.08 \\
Ours & 29.22 & 6.20 & \textbf{37.37} & \textbf{4.19}  & \textbf{32.99} & \textbf{7.81} & \textbf{56.94} & 20.09 \\
\textbf{$\bigtriangleup$ $(\uparrow)$} &
\textcolor[RGB]{0,0,0}{0.00} &
\textcolor[RGB]{34,120,5}{+2.07} &
\textcolor[RGB]{32,120,5}{+3.54} &
\textcolor[RGB]{34,120,5}{+3.59} &
\textcolor[RGB]{34,120,5}{+7.11} &
\textcolor[RGB]{34,120,5}{+1.56} &
\textcolor[RGB]{34,120,5}{+6.94} &
\textcolor[RGB]{34,120,5}{+2.01} \\
\bottomrule
\end{tabular}
\caption{
GUI grounding accuracy on ScreenSpot-Pro. 
All experiments are conducted under the same zero-shot prompt for fair comparison.}
\label{tab:gui_perception}
\end{table}

\section{Representative Case Studies}
\subsection{Mathematical Reasoning}

In this case study, we analyze two approaches to solving the problem of finding a list of positive integers with sum 30, unique mode 9, and a median that is a positive integer not in the list.

\begin{itemize}
    \item \textbf{GRPO Approach:} 
    The reasoning attempts to ensure 9 is the unique mode by including three 9’s. 
    While this satisfies the mode and sum conditions, it fails the median requirement: 
    the median is 9, which appears in the list. 
    As a result, although the calculation yields 248, the solution is invalid 
    because it violates the problem’s constraints. 
    This highlights how overlooking a single condition can compromise the correctness 
    of the entire solution.

    \item \textbf{CAPO Approach:} 
    The reasoning begins by deducing that the list must have an even number of elements, 
    since an odd-length list would force the median to appear in the list. 
    It then systematically constructs a 4-element list with two 9’s, 
    ensuring 9 remains the unique mode. 
    By carefully selecting the other two numbers as 5 and 7, 
    the solution achieves a median of 8, which is a positive integer not present in the list. 
    The final calculation yields a correct sum of squares of 236. 
    This approach demonstrates the effectiveness of structured, step-by-step analysis 
    in satisfying all given constraints.
\end{itemize}

\begin{figure}[!p]
    \centering
    \includegraphics[width=0.7\linewidth]{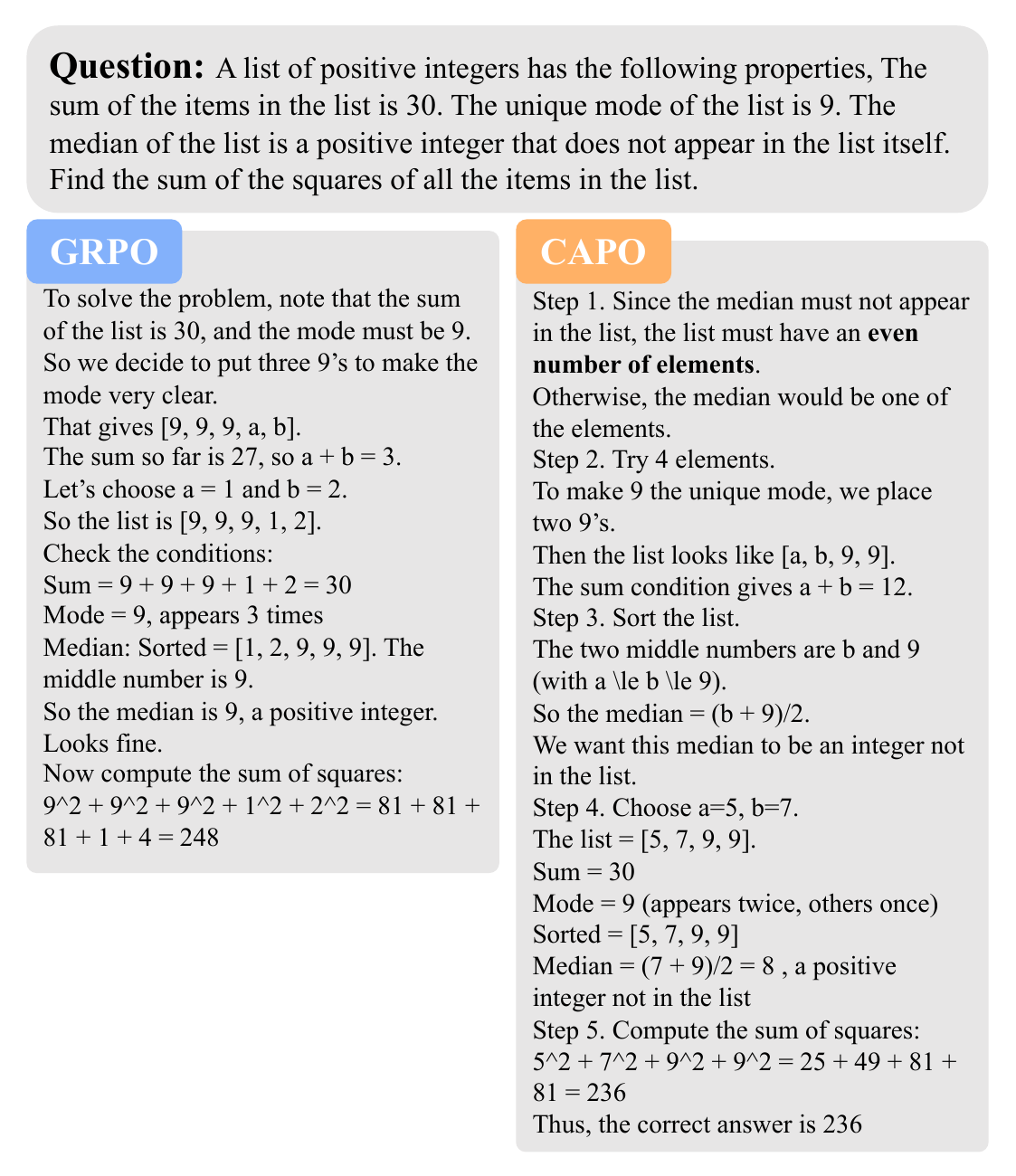}
    \caption{Comparison of two reasoning approaches for solving a number theory problem. The GRPO solution produces a flawed construction where the median condition is violated, leading to an incorrect result of 248. In contrast, the CAPO solution follows a structured step‑by‑step analysis, ensuring all conditions are satisfied, and derives the correct answer of 236.}
    \label{fig:placeholder}
\end{figure}

\subsection{Cross-Domain Case Studies on Graphical User Interface(GUI) Interaction}
In this case study, we analyze two approaches to following the instruction 
\textit{``find all items categorized under school supplies''} in a grocery shopping website interface.

This case study highlights the difference between exploratory, less targeted reasoning (GRPO) and structured, interface-aware reasoning (CAPO). While the GRPO strategy may appear reasonable, CAPO provides a principled and efficient path that ensures correctness and efficiency.

\begin{itemize}
    \item \textbf{GRPO Approach:} 
    The GRPO reasoning attempts to solve the task by searching for a category filter 
    that explicitly mentions ``School Supplies.'' 
    This approach is valid in principle but inefficient: 
    it assumes the user must scroll through the list of categories to locate the desired one, 
    without confirming whether it is already visible on the page. 
    As a result, the reasoning misses an immediate and more direct path to the goal.

    \item \textbf{CAPO Approach:} 
    The CAPO reasoning begins by explicitly restating the task goal, 
    then observes the visible interface for relevant cues. 
    It identifies that the category \textit{``School Supplies''} 
    is already present on the screen as a labeled button. 
    By proposing to click on this button, the approach directly navigates to the target category. 
    This method is both accurate and efficient, minimizing unnecessary exploration 
    and aligning closely with the ground truth action.
\end{itemize}

\begin{figure}[!p]
    \centering
    \includegraphics[width=0.7\linewidth]{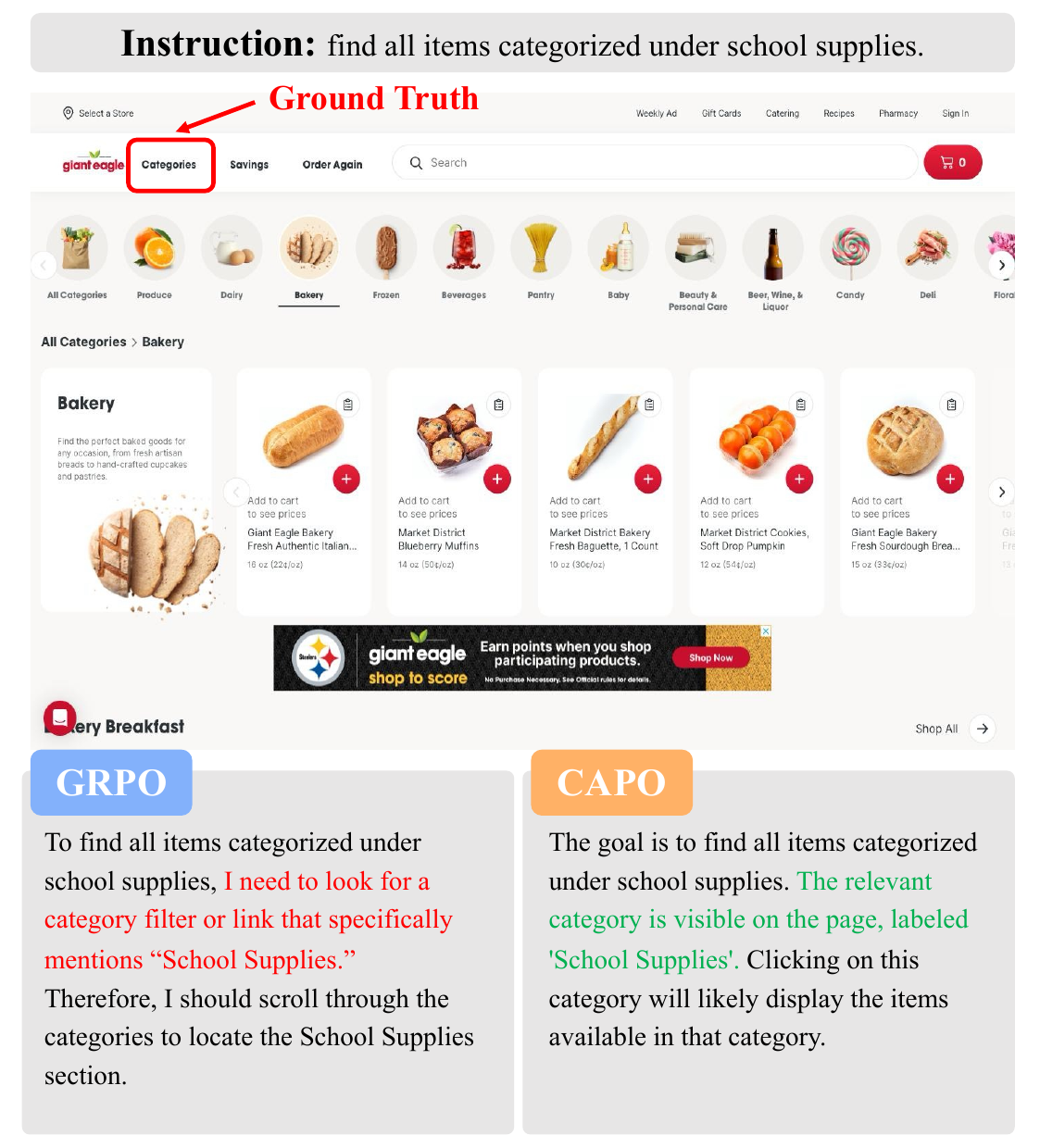}
    \caption{Comparison of two reasoning approaches for executing the instruction 
    \textit{``find all items categorized under school supplies''} in a web-based shopping interface. 
    The GRPO response relies on generic exploration without leveraging visible UI elements, 
    while the CAPO response explicitly identifies and utilizes the on-screen category filter, 
    leading to a more accurate and efficient solution.}
    \label{fig:placeholder}
\end{figure}

\end{document}